\documentclass{optica-article}

\journal{opticajournal} 

\articletype{Research Article}

\usepackage{mathtools}

\newcolumntype{P}[1]{>{\centering\arraybackslash}p{#1}}
\newcolumntype{M}[1]{>{\centering\arraybackslash}m{#1}}

\usepackage{tabularray}

\usepackage{xcolor}
\usepackage{changes}
\usepackage{graphicx}

\usepackage{float}

\begin{document}

\title{Universal Scale Laws for Colors and Patterns in Imagery}

\author{Rémi Michel\authormark{1,*} and Mohamed Tamaazousti\authormark{1}}

\address{\authormark{1}Université Paris-Saclay, CEA, List, F-91120, Palaiseau, France}

\email{\authormark{*}remi.michel@cea.fr} 


\begin{abstract*}
Distribution of colors and patterns in images is observed through cascades that adjust spatial resolution and dynamics. Cascades of colors reveal the emergent universal property that Fully Colored Images (FCIs) of natural scenes adhere to the debated continuous linear log-scale law (slope $-2.00 \pm 0.01$) (L1). Cascades of discrete $2 \times 2$ patterns are derived from pixel squares reductions onto the seven unlabeled rotation-free textures (0000, 0001, 0011, 0012, 0101, 0102, 0123). They exhibit an unparalleled universal entropy maximum of $1.74 \pm 0.013$ at some dynamics regardless of spatial scale (L2). Patterns also adhere to the Integral Fluctuation Theorem ($1.00 \pm 0.01$) (L3), pivotal in studies of chaotic systems. Images with fewer colors exhibit quadratic shift and bias from L1 and L3 but adhere to L2. Randomized Hilbert fractals FCIs better match the laws than basic-to-AI-based simulations. Those results are of interest in Neural Networks, out of equilibrium physics and spectral imagery.
\end{abstract*}

\section{Introduction}
Images (B\&W, RGB or multi-spectral) offer a never ending tricky composition of colored pixels \cite{burton1987color, ruderman1994statistics, srivastava2003advances, zontak2011internal, saremi2013hierarchical, roberts2022nature}. Their interpretation is of major interest in thematic imagery \cite{turcotte1997fractals, renosh2015scaling}, in statistical physics \cite{roggemann2018imaging, corberi2022many} and in image processing \cite{glasner2009super, zeiler2014visualizing}. 
Universal behaviors derived from images are of key interest because they allow to derive parameters of physical models and to put constrains on numerical models. Among them are the so-called scaling laws showing near fractal organisation of discrete patterns and the celebrated $1/f^\alpha$ densities in the continuous domain \cite{keshner19821,ruderman1997origins, chen2012zipf, bagrov2020multiscale}. Discrete patterns, textures, and local Hamiltonians of physical systems also exhibit universality, as seen with Potts models \cite{herpich2019universality} or with fluctuations of entropy across scales, which are notable in both image theory and physics \cite{jarzynski1997nonequilibrium, boccignone2000entropy, boccignone2001encoding, ferraro2002entropy, ferraro2004image, seifert2012stochastic, cocconi2022scaling}.
Numerous mathematical tools and theories in the field take benefit of those general behaviors, including Fourier descriptors, texture analysis, Wavelets \cite{wornell1993wavelet}, Local Binary Patterns \cite{ojala2002multiresolution, li2021multidimensional} and Markov random fields. These laws appear in the study of the deep neural networks which process those images \cite{alabdulmohsin2022revisiting}. They serve as valuable features for the data (including images) or for the internal representation of learning machines \cite{marsili2022quantifying}. They are more generally of interest in a variety of domains \cite{isherwood2017tuning, thurner2018introduction, kaplan2020scaling}.
But as of today, the origin of those scaling laws and the compliance of acquired images with them is still an open debate \cite{saremi2013hierarchical, wang2023origins}. In this paper we propose an investigation of universal color and spatial structures in images of natural scenes based on three main remarks.
The first remark is that patterns in images change with exposure time and sensitivity of recorders. This arises from the quantization and noise. A second remark is about the geometry of patterns. We basically choose the geometry of $2\times2$ square pattern because it's the most basic 2D pattern equally suitable for texture analysis, local Hamiltonian of physics and fractals. 
A latest remark is that patterns are conceptual. We thus choose the most basic rule to transform measurements using an unlabelled pattern descriptor that provides the same (resp. different) value at pixels which values $\sigma_i$ equal (resp. differ). Contrast invariant descriptors are unlabelled. Such unlabelled patterns also occur in statistical physics; the Ising (or Potts) Hamiltonian $H_i = -J \sum_{j \in V(i)} \delta (\sigma_i, \sigma_j)$ yields a 5 levels Hamiltonian  from unlabelled comparison between $\sigma_i$ and $\sigma_j$. The delta occurs from the Pauli exclusion principle in that case and we do not have such a principle in imagery. Thus we can not assign energy levels to patterns without assumptions. Thermodynamical models based on the assumption of the Ising models have been proposed that provide multi-scale descriptions of binarized representation of images \cite{stephens2013statistical, saremi2013hierarchical, saremi2014criticality, saremi2015wilson, saremi2015correlated, obuchi2016boltzmann, bagrov2020multiscale}. Following those studies, we propose to identify universal constants regarding colors and patterns in images of natural scenes. This will be based on a procedure that systematically captures spatial resolution and dynamics but without presuppositions about the energy levels of patterns.
  
In the following, we first present a database of images \textit{im} and introduce the Fully Colored Images. We then present a cascade \textit{C} derived from \textit{im} that allows to grasp the variations of patterns with dynamics and spatial scale. A local $2\times2$ 7-states texture patterns basis is derived from an Ising-like local Hamiltonian and the cascade is transformed into a cascade \textit{H} of the patterns describing local textures. 
Results describe statistics of cascades \textit{C} and \textit{H} through the notion of Shannon's entropy. 
The discussion encompasses power laws (continuous and discrete), the integral fluctuation of entropies with spatial scale and fractality through comparison with models.

\section{Method}
\subsection{Spectral Images} \label{sec:specIm}

Images used in this study are multi-channels images of various contexts, cropped by square of  $N\times N=256\times 256$ pixels, the number of spectral channels varies from 1 (B\&W images) to 32 and includes standard RGB 3-channels images (Table~\ref{Table 1: Database}). Typically all pixels  differ in images of natural scenes $256\times 256$ that have a dynamics of 1 octets or more and 3-5 or more  independent spectral channels; we refer to  them as Fully Colored Images (\textit{FCI}). RGB images used in this study include more than 97\%  of distinct colors (Table~\ref{Table 1: Database}). It's worth noting that any natural scene can potentially be captured as \textit{FCI}; it solely depends on the camera, including its dynamics and spectral channels.
\begin{table}
    \centering
    \label{Table 1: Database}
    \caption{Databases \cite{foster2006frequency, monno2015practical, nam2019real} of RGB and Multi-spectral Images ($256\times 256$ pixels). Percentage of independent colors typically close to 100\% (Fully Colored Images) for \{dynamics>256, Channels>3\}.}    \label{Table 1: Database}
    \begin{tabular}{c c c c c c}
        \hline
        Dyn & 256 & 256 & 256 & 300 & 300\\
        \hline
        Channels & 3 & 3 & 5 & 32 & 32\\
        \hline
        Colors (\%) & 94 & 95 & 100 & 100 & 100 \\
        \hline
        Ref & Baboon & Face & Flower & Foster1 & Foster2 \\
         \hline
        IMAGE & 
        \includegraphics[width=0.1\linewidth]{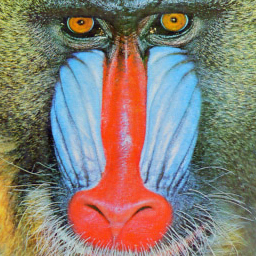} &
        \includegraphics[width=0.1\linewidth]{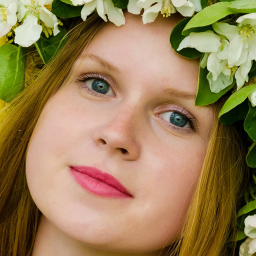} &
        \includegraphics[width=0.1\linewidth]{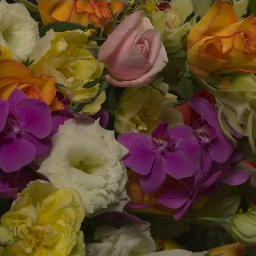} &
        \includegraphics[width=0.1\linewidth]{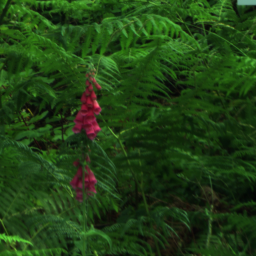} &
        \includegraphics[width=0.1\linewidth]{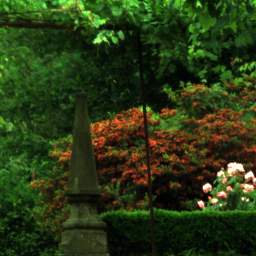} \\
        \hline
        \hline
        Dyn & 300 & 300 & 300 & 300 & 310 \\
        \hline
        Channels & 32 & 32 & 32 & 32 & 32 \\
        \hline
        Colors (\%) & 100 & 100 & 100 & 100 & 100 \\
        \hline
        Ref & Valley & Fern & Cuprite & Hyperion1 & Hyperion2 \\
         \hline
        IMAGE & 
        \includegraphics[width=0.1\linewidth]{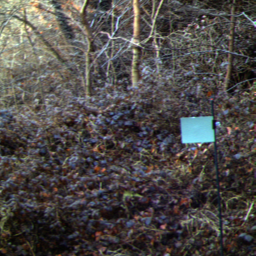} &
        \includegraphics[width=0.1\linewidth]{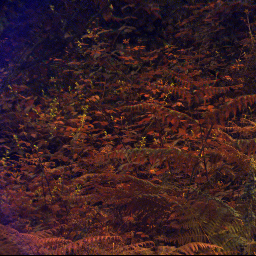} &
        \includegraphics[width=0.1\linewidth]{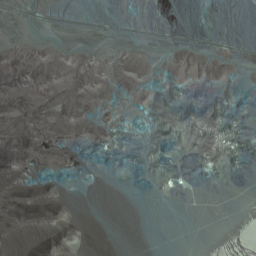} &
        \includegraphics[width=0.1\linewidth]{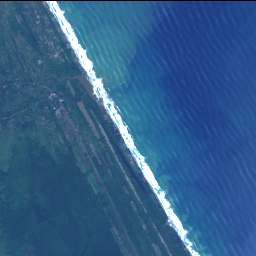} &
        \includegraphics[width=0.1\linewidth]{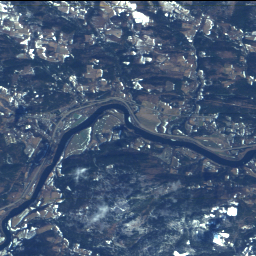} \\
        \hline
    \end{tabular}
\end{table}

\subsection{Cascade $C(k,s)$ of Spectral Images}

In order to grasp most information included in images we choose to analyse them at various spatial scales and dynamics through the cascade of images $C(k,s)$
\begin{equation}\label{eq:Cascadek}
    C(k,s)=[im*g(s)]_{\downarrow}/k
\end{equation}
where / denotes euclidean division, * denotes convolution, \textit{k} is integer in the range $[0,k_{\text{max}}=\max(\text{im}) + 1]$, \textit{g(s)} is a top at $s \times s$ normalized window and $\downarrow$ denotes sub-sampling by a factor $s$. $C(1,1)$ is \textit{im} and $C{(k_{\text{max}},s)}$ are black images, noted $\left[0\right]$ hereafter (see Fig.~\ref{fig 1:Image_exposure_time, cascade k}).
\begin{figure}[H]
\centering
\includegraphics[width=1\linewidth]{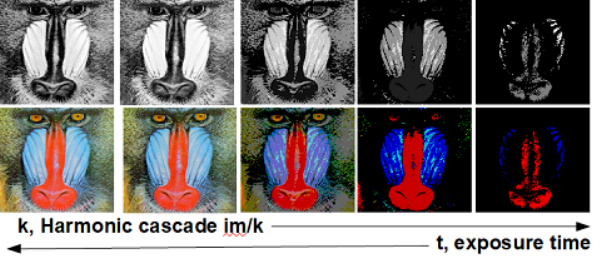}
\caption{ Bottom: Patterns appear during exposure time,  cascade $C(k,1)=im/k$ .Top : flag of independent colors in levels of grey. Bottom : im/k, rescaled to 1 octet }
\label{fig 1:Image_exposure_time, cascade k}
\end{figure}

\subsection{Local $2\times2$ Hamiltonian and Patterns}

At point \textit{M} of an image\textit{ im}, we define the local texture from the hamiltonian $H_{\text{im}}(M)$:
\begin{equation}\label{eq:UN}
H_{im}(M) = \sum_{(i, j) \in V(M)} \frac{\delta(\sigma_i, \sigma_j)}{d(i,j)}
\end{equation}
where \textit{V}\textit{ }is the $2\times 2$ square including \textit{M} at bottom left and \textit{d(i,j)} the euclidean distance. $H_{\text{im}}(M)$ is both unlabelled, through the comparison made by the $\delta$ function, and invariant per rotation of the 4 pixels in \textit{V(M)}. For each loop of 4 colored (spectral) pixels $H_{\text{im}}(M)$ can take 7 distinct values. $H_{\text{im}}(M)$  reduces each spectral loop to its class of equivalence in the set of the 7 unlabelled necklaces that have minimum lexicographic representatives {0000, 0001, 0011, 0012, 0101, 0102, 0123}. $H_{\text{im}}(M)$ thus describes the local structure (or local texture) and those 7 patterns constitute the basis of patterns of the study (see Table~\ref{Table 2: Unlabelled Necklaces,Hamiltonian}).

\begin{table}[H]
    \centering
      \label{Table 2: Unlabelled Necklaces,Hamiltonian}
      \caption{Local Patterns Basis. Each local loop of four pixels yields one of the 7 values $H_i$ of the Hamiltonian representing the class of equivalence in the set of the 7 unlabeled necklaces $(4,4)$.}
    \label{Table 2: Unlabelled Necklaces,Hamiltonian}
    \begin{tabular}{m{2cm} m{1cm} m{1cm} m{1cm} m{1cm} m{1cm} m{1cm} m{1cm}} 
        \hline
        Local Pattern  &        
        \begin{center}\includegraphics[width=0.95\linewidth]{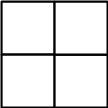} \end{center}&
        \includegraphics[width=1\linewidth]{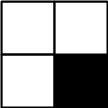} &
        \includegraphics[width=1\linewidth]{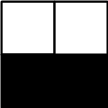} &
        \includegraphics[width=1\linewidth]{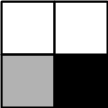} &
        \includegraphics[width=1\linewidth]{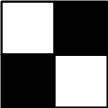} &
        \includegraphics[width=1\linewidth]{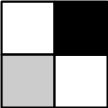} &
        \includegraphics[width=1\linewidth]{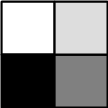} \\                
        \hline
        \centering Unlabeled Necklace & \centering  0000 & \centering 0001 & \centering 0011 & \centering 0012 & \centering 0101 & \centering 0102 &  \:0123 \\
        \hline
    \end{tabular}
\end{table}

\begin{figure}[H]
    \centering
    \includegraphics[width=0.95\linewidth]{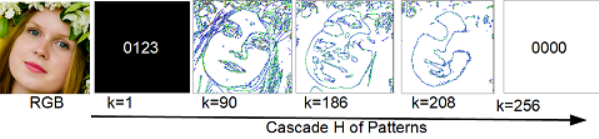}
    \caption{Patterns evolution with \(k\), cascade \(C(k,1)=\frac{im}{k}\) for the 7 patterns (arbitrary colors). At \(k=1\) in most images local patterns resume to 0123 because all local  \(\sigma_i\) differ. For \(k>\max(\text{im})\), $C(k,1)=\left[0\right]$ and patterns resume to 0000. In between, patterns fluctuate with universal maximum $1.74\pm 0.013$ for Shannon entropy and at compliance with the Integral Fluctuation Theorem for entropy production over scales (see text for details). Original image (left) from database \cite{nam2019real}.}
    \label{fig 2:PatternsEvolutionWithK}
\end{figure}

$H_{C(1,s)}$ nearly equals $\left[0\right]$ because high spatial frequency variations of $\sigma_i$ in images reduces most spectral loops $2\times 2$ to pattern 0123 , $H_{FCI}=\left[0\right]$ because all $\sigma_i$ differ and $H_{C{(k_{\text{max}}},s)} = 4 + 2 \sqrt{2}$ because the null image always reduces to loops equivalent to 0000 (see Fig.~\ref{fig 2:PatternsEvolutionWithK}).
\section{Results}
\subsection{Log-scale Entropy of Fully Colored Images}

\begin{figure}[H]
\centering
\includegraphics[width=1\linewidth]{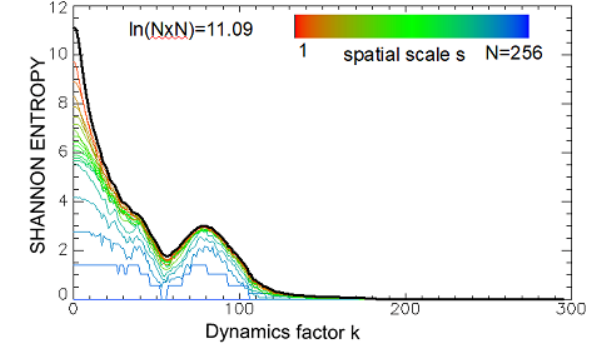}
\caption{Entropy $S_{C}$ of $C(k,s)$. Cuprite image $(256\times 256)\times 16$ channels, Fully Colored. $S_{C_k}$ depends only on  density of states of $C(1,s)$ by binning into factor\textit{ k}. It does not decrease monotonically with \textit{k} which contributes to variations in patterns along the cascade.}
\label{fig3 Entropy_Image_k}
\end{figure}

\begin{figure}[H]
\centering
\includegraphics[width=1\linewidth]{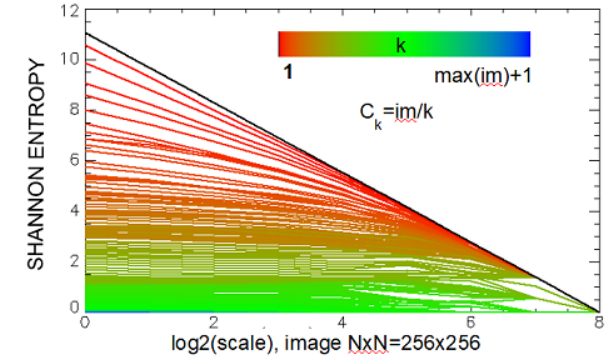}
\caption{Entropy $S_C$ of the cascade \textit{(k,s)}. Hyperion image $(256\times 256)\times 32$ channels, Fully Colored. The linear Law is universal for Fully Colored Images (black line). A lower entropy, curvature at high resolution (low $s$) and bias at low resolution (high $s$).}
\label{fig4 Entropy_Image_cascade_scale}
\end{figure}

\begin{figure}[H]
\centering
\includegraphics[width=1\linewidth]{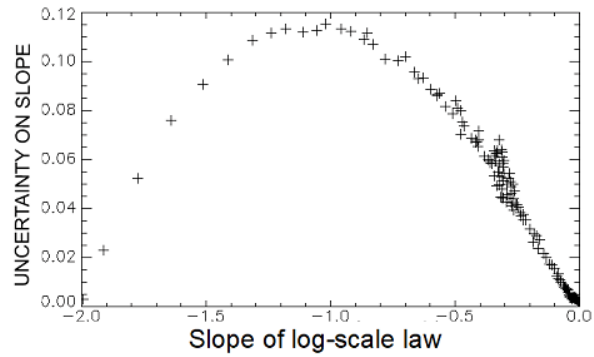}
\caption{Shift to $\log(s)$ as a Entropy of $C(k,s=1)$ with k. Cuprite image $(256\times 256)\times 8$, Fully Colored. FCI images (bottom left), are perfectly linear with slope $a=-2$ while best linear match at lower entropy of $im$ reveals higher order contributions, when the image in near zero, both $a$ and its standard deviation tends to zero.}
\label{fig5 Entropy_Image_scale_sigma}
\end{figure}

The Shannon entropy of images is defined as:
\begin{equation}
S_{im} = -\sum_{m} p(m) log(p(m))    
\end{equation}
where \textit{m }is the number of different $\sigma_i$ in the image and \textit{p(m)} is the probability associated with \textit{$\sigma_i$}. The entropy of the cascade $S_{C}$ for varying values of \textit{k} and \textit{s} are presented in Fig.~\ref{fig3 Entropy_Image_k}. The density of states (id. the histogram) of  ${C(k,s)}$ is derived from that of  $C(1,s)$  by binning the density of states by a step \textit{k}. The entropy decreases with\textit{ k}, though not monotonically, is maximum for\textit{ k}=1 and zero at $ k_{max}$. 
Variations of entropies with scale  \textit{s } depends on the organisation of pixels in the image, The entropies vary in a near linear log-scale law  (Fig.~\ref{fig4 Entropy_Image_cascade_scale}). The greater the Shannon entropy of \textit{im}, the more linear the decrease. The variations of the entropy  with scale \textit{s} shows that \textit{FCI }have a universal log-linear variation with spatial scale \textit{s}:
\begin{equation}\label{FCI_ENTROPY}
S_{FCI}(s) = -2 \log\left(\frac{s}{N}\right) \pm 0.01
\end{equation},
thus validating for \textit{FCI} the $1/f^2$ law \cite{Schaaf1996ModellingTP}. When the image is not \textit{FCI} than standard deviations on the estimates of the two coefficients of the best least-square linear fit increases, (see Fig.~\ref{fig5 Entropy_Image_scale_sigma}).

\subsection{Entropy of patterns}
\begin{figure}[H]
\centering
\includegraphics[width=1\linewidth]{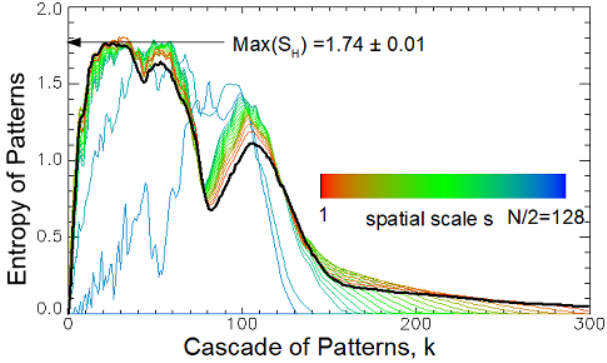}
\caption{Entropy of $H{(C)}$, 7 states. Universal Maximum of Entropy of patterns $S_H$=1.74 $\pm$0.013, constant with scales for Fully Colored Images, below the maximum $\ln(7)$. Hyperion image $(256\times 256)\times 32$ channels, Fully Colored (Hyperion 1)}.
\label{fig6 Entropy_UN_k}
\end{figure}

\begin{figure}[H]
\centering
\includegraphics[width=1\linewidth]{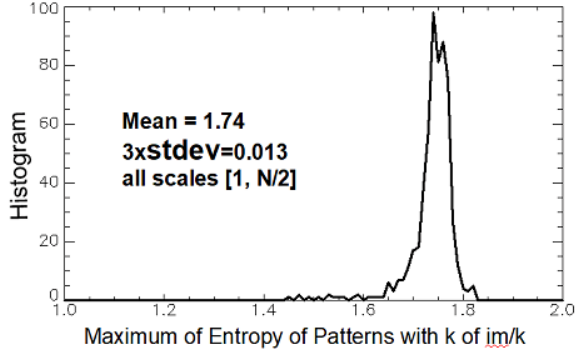}
\caption{Maximum of Entropy of $H({C})$ over database. Universal Maximum of entropy $S_H$=1.74 $\pm$0.01, constant over scale for Fully Colored Images. Below $\ln(7)$, this maximum is interpreted as resulting from the scarcity of patterns 0101 and 0102 in images of natural scenes.}
\label{fig7 MAXI_Entropy_UN_k}
\end{figure}

For any image $S_{H\left[{C(k_{max},s)}\right]}=0$ because $C(k_{max},s)=\left[0\right]$, for most images $S_{H\left[{C{(1,1)}}\right]}$ is closed to zero as pattern 0123 dominates in textured images and $S_{H{\left[FCI\right]}}=0$. In between, for varying values of \textit{k }in the harmonics cascade, $S_{H{(FCI)}}$ presents an universal maximum equals to $1.74\pm 0.013$ at each spatial scale, below the maximum $\ln(7)=1.93$ (see Fig.~\ref{fig6 Entropy_UN_k} and Fig.~\ref{fig7 MAXI_Entropy_UN_k}).
\begin{equation}\label{MAX_ENTROPY_FCI}
\max_{k} \left[S_{H{(FCI)}}\right] = 1.74 \pm 0.013
\end{equation}
\begin{equation}\label{FCI_derive ENTROPY}
\frac{\partial}{\partial s} \max_{k} \left[S_{H{(FCI)}}\right] = 0
\end{equation}

\begin{figure}[H]
\centering
\includegraphics[width=1\linewidth]{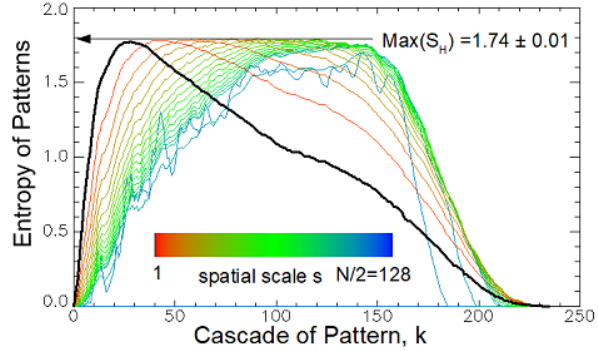}
\caption{Entropy of $H{(C)}$. The production of Entropy varies smoothly with scale (from black, red to deep blue) at depends on k. Flower image $(256\times 256)\times 5$ channels, Fully Colored.} .
\label{fig8 Entropy_UN_k}
\end{figure}

\begin{figure}[H]
\centering
\includegraphics[width=1\linewidth]{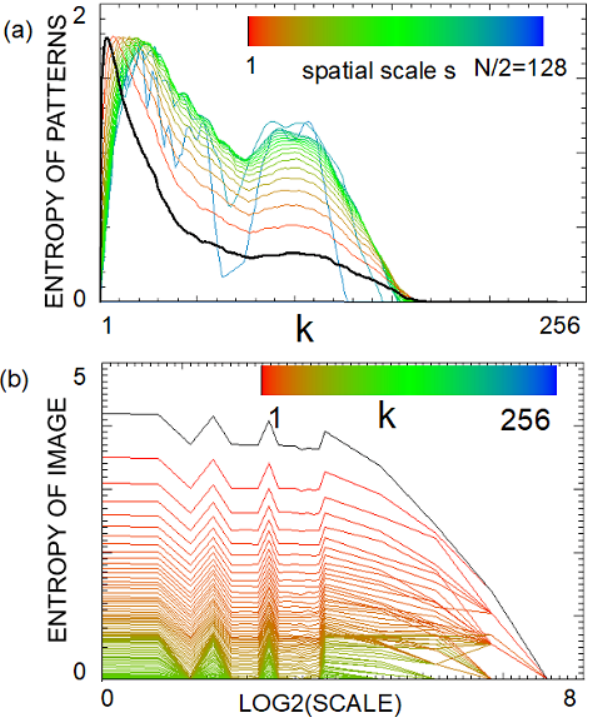}
\caption{Non \textit{FCI images of Natural Scene (Flower image ($256 \times 256 \times 2$) channels)} tend to maintain $max_{k}(S_{H})=1.74$ though they do not respect log-scale linearity.}
\label{fig9 Entropy_UN_k_NONFULLY}
\end{figure}

The entropy production varies smoothly with scales along the cascade at rates changing with the level \textit{k} (see Fig.~\ref{fig8 Entropy_UN_k}). Those variations resemble that reported for the complex dynamics of physical systems such as turbulent cascade in fluid dynamics, a domain where entropy production and transfer with scale is of paramount importance \cite{seifert2012stochastic, alexakis2018cascades}. 
A trade-off between $\log(s)$ law and patterns distribution is noticeable in images that are far from being \textit{FCI } (see Fig.~\ref{fig9 Entropy_UN_k_NONFULLY}). In that case, important shifts in the $\log(s)$ law occur while the maximum of the entropy of pattern is maintain. 

\subsection{Entropy Production and Integral Fluctuation Theorem for Natural Scene}

\begin{figure}[H]
\centering
\includegraphics[width=1\linewidth]{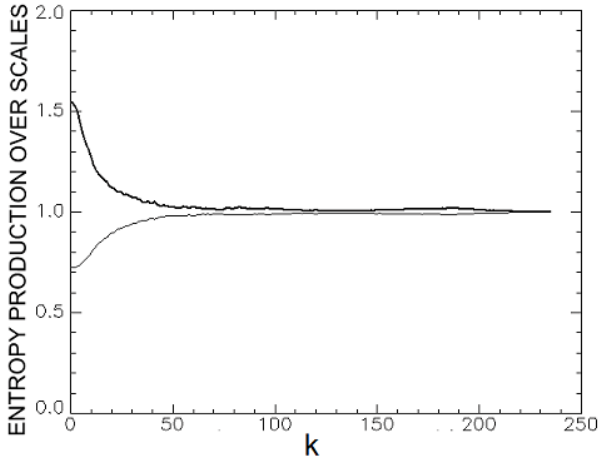}
\caption{Entropy Production of  Image with scale of $C$ (direct bold). Entropy production is balanced (grey) but does not respect the Integral Fluctuation Theorem. Hyperion image $(256\times 256)\times 32$ channels, Fully Colored.}
\label{fig10 Entropy_Production_Image}
\end{figure}

\begin{figure}[H]
\centering
\includegraphics[width=1\linewidth]{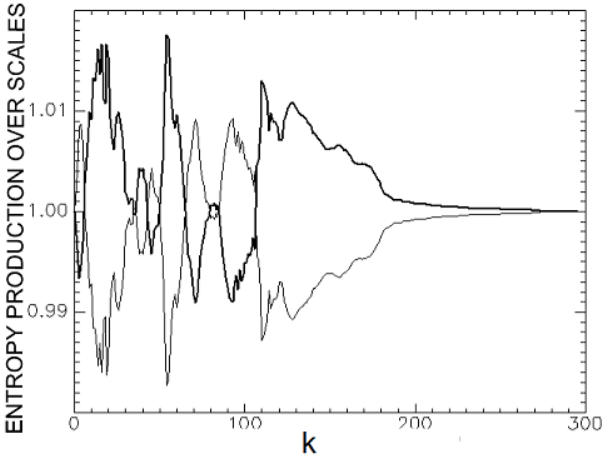}
\caption{Entropy Production with scale of Patterns follows the celebrated Integral Fluctuation Theorem (\textit{IFT}) within $\pm 0.01$ for all Natural Scenes. At some values of \textit{k}, exact \textit{IFT} is achieved. Cuprite image $(256\times 256)\times 16$ channels, Fully Colored.}
\label{fig11 Entropy_Production}
\end{figure}

The total production of entropy $\Omega_{\pm}(im)$ as a function of the spatial scale \textit{s}  (\textit{s} being an integer) is defined as
\begin{equation}\label{IFT}
\Omega_{\pm}(im)=\frac{1}{(N-1)}\sum_{s=1}^{N-1}  \exp{\pm\Delta S_{im}(s+1,s)}
\end{equation}
$\Delta S_{im}(s+1,s)$ is the difference of Shannon entropies $S_{im}$ at spatial scales \textit{s}+1 and\textit{ s}. $\Omega_{\pm}(C_k)$ fluctuates over \textit{k} with values typically in the range $[0.5 , 1.8]$ (see Fig.~\ref{fig10 Entropy_Production_Image}).
Because the pattern is $2\times 2$, the production $\Omega_{\pm}(H)$ of the Hamiltonian \textit{H} describing local structures  is limited in scale to \textit{N}/2:
\begin{equation}\label{IFT_UN}
\Omega_{\pm}(H)=\frac{1}{(N/2-1)}\sum_{s=1}^{N/2} \exp{\pm\Delta S_H(s+1,s)}
\end{equation}
For \textit{FCI} we get, for all values of\textit{ k} in the harmonic cascade (see Fig.~\ref{fig11 Entropy_Production}):
\begin{equation}\label{IFT_UN_FCI}
\Omega_{\pm}(H)=1.00\pm0.01
\end{equation}
Those values respect the so-called Integral Fluctuation Theorem and denote a hallmark of chaotic cascade that behave far from  equilibrium. This property holds for all values of \textit{k} in the cascade of \textit{FCI} (see Fig.~\ref{fig11 Entropy_Production}). At some values of \textit{k} , $\Omega_{\pm}(H)=1.00\pm1e^{-6}$, images presents balanced equilibrium with the maximum achievable accuracy in our context.

\begin{figure}[H]
\centering
\includegraphics[width=1\linewidth]{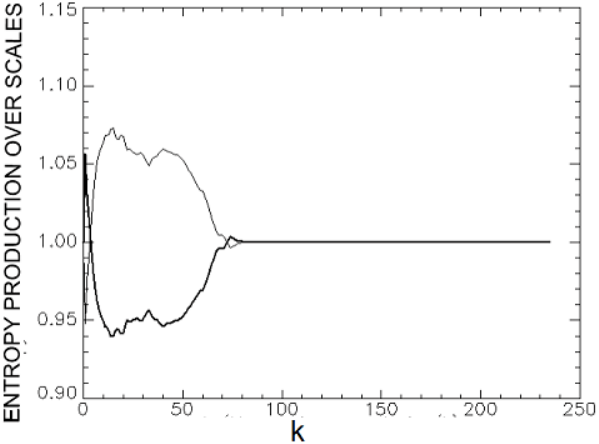}
\caption{Entropy Production of Patterns with scale in images of natural scenes far from being FCI is maintains. Flower image $(256 \times 256) \times 2$ channels.}
\label{fig12 Entropy_Production_NONFULLY}
\end{figure}

When the images are not \textit{FCI}, the values of the entropy production with scale do not shift from 1 with \textit{k} (Fig.~\ref{fig12 Entropy_Production_NONFULLY}). supporting the assumption that the hamiltonian maintains the complexity of patterns with scale despite the log-scale law does not hold anymore.

\section{Discussion}

Simulating these three behaviors is far from a straightforward procedure. Such a complexity is particularly evident in domains dealing with intricate 2D (or higher-dimensional) systems. In the realm of image-oriented Artificial Intelligence, the convergence of super-resolution and the generation of fake images produces visually realistic and intricate patterns, making them ideal candidates for testing the proposed framework. Fields of physical phenomena that exhibit these behaviors include turbulence in fluid mechanics, non-equilibrium thermodynamics, biology, vision, finance and imagery \cite{martyushev2006maximum, renosh2015scaling, kondi2023using}. In these domains, an history of studies has demonstrated the coherence between concepts such as fractal behavior, pattern orientation, the $1/f^\alpha$ spectral response of continuous models and Self-Organized Criticality  \cite{bak1988self, beggs2003neuronal, chen2012zipf}. The following simulations are inspired by those studies and aim at best reproduce the described behavior of images \cite{pesquet2002stochastic}.

\subsection{About the Scarcity of Some Patterns in Natural Scenes}
\begin{figure}[H]
\centering
\includegraphics[width=1\linewidth]{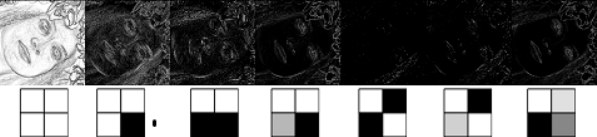}
\caption{Integral of each pattern with k. s=1, shows that 0101 and 0102 are the least present pattern in the field: \textbf{Relative Abundance (\%) : 71.77, 12.00, 7.09, 4.32, 0.55, 1.34, 2.92}.}
\label{fig13 Scarcity}
\end{figure}

The 7 patterns of this study, the Unlabelled Necklace (4,4), do not appear equally and pattern 0101 can hardly be found in images as shown with the integral of each pattern with \textit{k} in the harmonic cascade (Fig.~\ref{fig13 Scarcity}).

We suggest that this scarcity results from the smoothness of natural patterns while information about smoothness is partially lost in the basic $\delta$ function of the hamiltonian of Eq.~\eqref{eq:UN}. A tile of the elevation of planet Earth as a single channel image shows high variations in the relative abundance of the 7 patterns; the scarcest, 0101, represents only 0.36\% of patterns of the topographic tile and so mountain pass is the scarcest pattern in elevation at any scale (Table~\ref{Table 3: RELIEF_SCARCITY}). This scarcity give insight why $max_{k}\left[S_{H({\text{Plan}})}\right]< ln(7)$. The values $H_i$ of the 7 necklaces from the hamiltonian of Eq.~\eqref{eq:UN} do not provide direct incentive on how a physical model (e.g. a thermodynamical model with Boltzmann-like statistics) could recover this particular property at this step; adding an interaction with a to-be-defined external field that acts distinctly with the $\sigma_i$ or other kind of models, like Self-Organized-Criticality may further help to describe results obtained in Table~\ref{Table 3: RELIEF_SCARCITY}.
\begin{table}[H]
    \centering   
    \caption{Relative abundance of patterns in a Tile of the elevation of Earth (extracted from gtopo30 \cite{GTOPO30}) shows that necklace 0101 is the scarcest.}
    \label{Table 3: RELIEF_SCARCITY}
    \begin{tabular}{c c c c c c c c}
        \hline
        Pattern & 0000 & 0001 & 0011 & 0012 & 0101 & 0102 & 0123 \\
        \hline
        Occurrence (\%) & 80.9 & 8.1 & 4.7 & 2.9 & 0.36 & 0.96 & 2.0 \\
        \hline
    \end{tabular}
\end{table}

\subsection{Simulations}
\begin{figure}[H]
\centering
\includegraphics[width=0.75\linewidth]{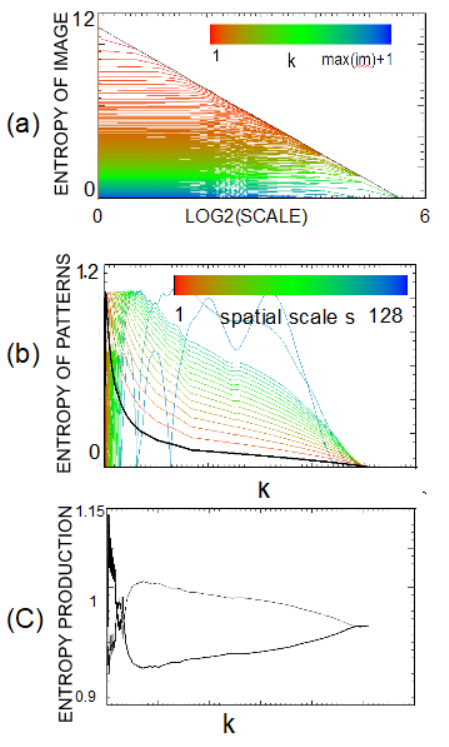}
\caption{Plane FCI ($256 \times 256 \times 2$ channels) respects log-scale (a)  and balanced equilibrium (c) but yields $max_{k}\left[S_{H({\text{Plan}})}\right] = 1.05$. (b)}
\label{fig14 PLAN_3_laws}
\end{figure}

\begin{figure}[H]
\centering
\includegraphics[width=0.75\linewidth]{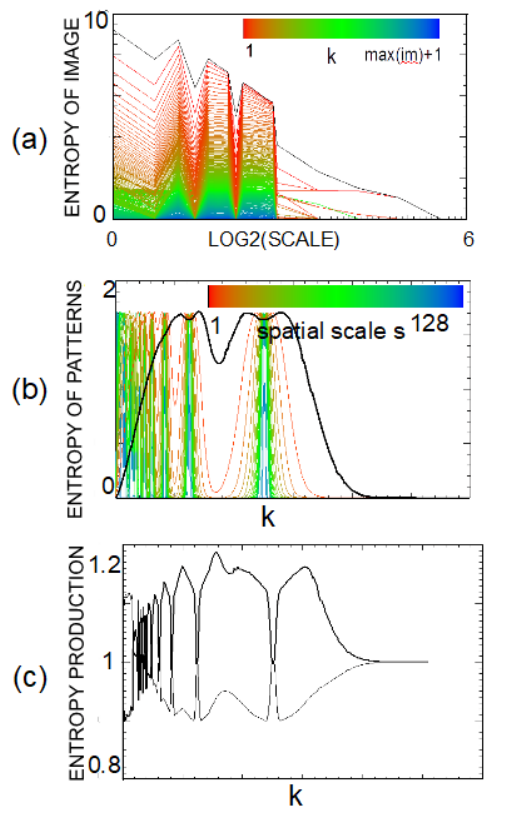}
\caption{Random FCI ($256\times 256\times 2$ channels) does not respect any of the 3 rules, 
 log-scale (a) ,$max_{k}\left[S_{H({\text{Random}})}\right] = 1.70$ (b) and poorly balanced equilibrium (c)  (see text).}
\label{fig15 Random_3_laws}
\end{figure}

It's easy to construct images that present $S_H$ greater than 1.74 and even equal to $\ln(7)$, just paving regularly the plane equally with the 7 patterns, those images just do not appear in natural scenes Table~\ref{Table4: Image ln(7)}.
\begin{table}
    \centering
    \caption{Example of image with \(S(H) = \ln 7\). First two lines repeatable \textit{ad libitum}, new mirror columns may be added easily with different labels; all seven patterns always appear equally. This situation does not appear in natural scenes.}
    \label{Table4: Image ln(7)}
    \begin{tabular}{c c c c c c c c }
        \hline
        0 & 0 & 0 & 1 & 1 & 0 & 0 & 2\\
        \hline
        0 & 0 & 1 & 0 & 0 & 2 & 1 & 3\\
        \hline
        \textit{0} & \textit{0} & \textit{0} & \textit{1} & \textit{1} & \textit{0} & \textit{0} & \textit{2}\\
        \hline
        \textit{0} & \textit{0} & \textit{1} & \textit{0} & \textit{0} & \textit{2} & \textit{1} & \textit{3}\\
        \hline
    \end{tabular}
\end{table}
Let's \textit{im} be the most basic\textit{ FCI} plane $im(c,l)=c+l*N$. In order to reduce the intractable dynamics $N^2$ of \textit{im} and to mimic the correlation between spectral channels of images of natural scenes, we  map  monochromatic $N\times N$ images on the 2 channels \textit{FCI} image with dynamics 2\textit{N}, $im(c,l,0)=c+l$ and $im(c,l,1)=N+c-l$. Results are presented in Fig.~\ref{fig14 PLAN_3_laws}, this trivial FCI image respects the log-scale law but the  value of the maximum of entropy, though independent on the scale, only equals to  
\begin{equation}\label{MAX_ENTROPY_Plan}
max_{k}\left[S_{H({\text{Plan}})}\right] = 1.05 \pm 0.01
\end{equation}
Such result was expected because of the low "complexity" of the plane but that case is still noticeable because it is opposite to that of Fig.~\ref{fig9 Entropy_UN_k_NONFULLY} of a non fully colored natural scene that manage to respect $max_{k}\left[S_{H({\text{Plan}})}\right] = 1.74 \pm 0.01$ though not respecting the $\log(s)$ law. 
Hence we may infer that at low energy (high values of k in the harmonic cascade $im/k$) the image maintains the complexity of its organisation in term of patterns but not the log-scale law while basics models of \textit{FCI} do the opposite.
The same procedure has been used with a fully colored Random image \textit{(N,N,2)} with dynamic \textit{N} that includes all the 2 spectrum located randomly. None of the three observations is respected, the log-scale law is highly biased, the maximum of entropy of pattern is 1.70 $\pm0.01$ and no balanced fluctuations (see Fig.~\ref{fig15 Random_3_laws}).

\begin{figure}[H]
\centering
\includegraphics[width=1\linewidth]{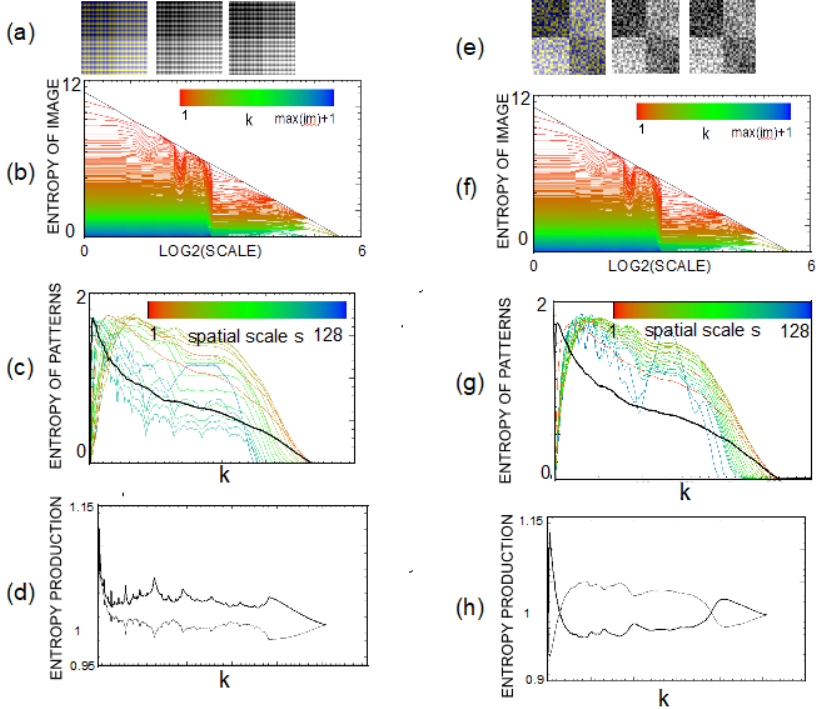}
\caption{Modified Hilbert Fractal FCI ($256\times 256\times 2$ channels, dynamics 2*N) best fit the three laws. Randomness in the angle of rotation of the Hilbert $2\times 2$ 0123 pattern (top) allows for a better fit of pattern entropy and a more accurately balanced entropy production than regular pattern (bottom). snapshots: Image 2 channels (Left) green-red, (Middle) channel1, (Right) channel2.}
\label{fig16 Fractal_3_laws}
\end{figure}

Computer generated images \cite{dong2015image} propose super-resolved images that include to few entropy at extrapolated scales $\times 2$ and stand below the log-scale law at interpolated scales s. Fakes build on statistically "melting" real faces better manage to respect all the three laws. The face image (see Table~\ref{Table 1: Database}) reduced to $64\times 64$ pixels and than super-resolved by a factor 4 by  gets entropies at super-resolved scales $s=1$ and $s=2$ that are respectively 20\% and 5\% below the expected values for \textit{FCI}, showing that the AI does not recover all of the details of the image \cite{deepai}. A major inconvenient of those AI generated images for this study is that the process mainly transfer the complexity of images we want to grasp into the  network (during training procedures). The neural networks may, when considered as graphs,  exhibit similar properties than those described in this study  especially when they are trained using images of natural scenes. Thus similar asymptotic laws may hold in the graphs which may be of help to better constrain their design. 
Fractal distribution of pattern has been studied and proposed for modeling of stochastic 2D process showing the relationship between $1/f$ and fractals \cite{chen2012zipf}. In the context of the proposed $2 \times 2$ Hamiltonian and the requirement of a \textit{FCI}, the Hilbert fractal of the $2\times 2$ pattern 0123 allows to provide a Fully Colored Image that presents the pattern 0123 everywhere at any spatial scale s: At step 0 the square image is divided in 4 $N/2 \times N/2$ sub-squares that are attributed a $2\times2$ permutation $\sigma_i^0$ of the vector $[0,1,2,3]$. The iterative process in \textit{j} consist in replacing each square $\sigma_i^j$ by four sub-squares with new values:
\begin{equation}\label{Hilbert}
\{\sigma_i^{j+1}\}=4*\sigma_i^j+P_j
\end{equation}
where $\{.\}$ denotes the set of 4 news values and $P_j$ a permutation of [0,1,2,3]. During the iterative process building the fractal we choose randomly the permutation $P_j$ to get a randomized Hilbert fractal \textit{f}. The fractal includes all integers between 0 and $N^2-1$ by construction. It is thus a \textit{FCI}  at any spatial scale s. The image is then mapped on a 2 channels Fully Colored Image of dynamics \textit{2N}: 
\begin{equation}\label{Hilbert_channels}
Fractal(i,0) = \frac{f(i)}{N} + \text{f}(i) \mod N
\end{equation}
\begin{equation}
Fractal(i,1) = N + \frac{f(i)}{N} - \text{f}(i) \mod N
\end{equation}
Modified Hilbert Fractals FCI ($256\times 256\times 2$ channels, dynamics $2N$) best fit the three laws. Randomness in the angle of rotation of the Hilbert $2\times 2$ 0123 pattern allows for a better fit of pattern entropy and a more accurately balanced entropy production, see Fig.~\ref{fig16 Fractal_3_laws}. 

\section{Conclusion}
Patterns can be distinguished, ordered and eventually valued. General studies of natural images relying on analogies to physical model assumes patterns with assigned values. Based on the weakest assumption of distinguishable patterns, we observe in this study three universal laws of natural images associated to scales and dynamics. This work may inspire advancements in neural networks-based image analysis. In particular, the design and training of these networks may benefit from the three reported universal observations about images of natural scenes. The study also offers a context of physical complex phenomena, as log-scale laws (L1) and entropy fluctuation (L3) in images have analogs in physics of chaotic systems. We did not find in physics an equivalent of the law L2 about an universal value of the entropy, reached at some dynamics, of a local Hamiltonian of interaction. We propose that optimal settings of recorders should offer dynamics and channels capable of capturing fully colored images of  studied scenes. Furthermore, this study provides a combinatorial perspective on natural scenes, depicting them as sets of fully colored tiles adhering to three universal multi-scale combinatorial constraints regarding their partitioning into the 2D puzzle-image.

\newpage
\begin{backmatter}

\bmsection{Acknowledgments}
This study benefited from short yet invaluable discussions with Oriol Bohigas, Christian Lavault and Henning Bruhn-Fujimoto.

\bmsection{Data availability} Data underlying the results presented in this paper are available in Ref. \cite{foster2006frequency, monno2015practical, nam2019real, GTOPO30, deepai}.

\end{backmatter}

\bibliographystyle{unsrt}
\bibliography{sample.bib}

\end{document}